\documentclass[runningheads]{llncs}
\usepackage{graphicx}
\usepackage{epsfig} 
\usepackage{mathptmx} 
\usepackage{times} 
\usepackage{amsmath} 
\usepackage{amssymb}  

\usepackage{hyperref}
\usepackage{authblk}
\usepackage{url}
\usepackage{todonotes}
\usepackage{subcaption}
\usepackage{stmaryrd} 
\usepackage{breakcites}

\usepackage{authblk}
\usepackage{adjustbox}

\begin{document}
\title{\LARGE \bf
Training Discriminative Models to Evaluate Generative Ones
}
\titlerunning{Training Discr. Models to Evaluate Generative Ones}

\author{Timoth\'ee Lesort\inst{1, 2} \orcidID{0000-0002-8669-0764} \and
Andrei Stoian\inst{2} \orcidID{0000-0002-3479-9565} \and
Jean-Fran\c{c}ois Goudou\inst{2} \and
David Filliat\inst{1} \orcidID{0000-0002-5739-1618} }

\authorrunning{T. Lesort et al.}

\institute{Flowers Laboratory (ENSTA ParisTech \& INRIA), France \and
 Thales, Theresis Laboratory, France}

\maketitle              
\begin{abstract}

Generative models are known to be difficult to assess. Recent works, especially on generative adversarial networks (GANs), produce good visual samples of varied categories of images. However, the validation of their quality is still difficult to define and there is no existing agreement on the best evaluation process. This paper aims at making a step toward an objective evaluation process for generative models. It presents a new method to assess a trained generative model by evaluating 
the test accuracy of a classifier trained with generated data. 
The test set is composed of real images.
Therefore, The classifier accuracy is used as a proxy to evaluate if the generative model fit the true data distribution.
By comparing results with different generated datasets we are able to classify and compare generative models. 
The motivation of this approach is also to evaluate if generative models can help discriminative neural networks to learn, i.e., measure if training on generated data is able to make a model successful at testing on real settings.
Our experiments compare  different generators from the Variational Auto-Encoders (VAE) and Generative Adversarial Network (GAN) frameworks on MNIST and fashion MNIST datasets.
Our results show that none of the generative models is able to replace completely true data to train a discriminative model. But they also show that the initial GAN and WGAN are the best choices to generate on MNIST database (Modified National Institute of Standards and Technology database) and fashion MNIST database.

\end{abstract}

\section{Introduction}
\label{introduction}

Generative models are machine learning models that learn to reproduce training data and to generalize it. This kind of model has several advantages, for example as shown in \cite{ng2002discriminative}, the generalization capacity of generative models can help a discriminative model to learn by regularizing it.
Moreover, once  trained, they can be sampled as much as needed to produce new datasets. Generative models such as GAN (Generative Adversarial Network) \cite{Goodfellow14}, WGAN (Wasserstein-GAN)\cite{Arjovsky17}, CGAN (Conditional-GAN) \cite{Mirza14}, VAE (Variational Auto-Encoder) \cite{Kingma13} and CVAE (Conditional-VAE) \cite{Sohn15} have produced samples with good visual quality on various image datasets such as MNIST, bedrooms \cite{Radford15} or imageNet \cite{Nguyen16}.
They can also be used for data augmentation \cite{Ratner17}, for safety against adversarial example \cite{Wang17}, or to produce labeled data \cite{Sixt16} in order to improve the training of discriminative models. 

One commonly accepted tool to evaluate a generative model trained on images is visual assessment. It aims at validating the realism of samples. However those methods are very subjective and dependent on the evaluation process.  One case of this method is called ``visual Turing tests'', in which samples are visualized by humans who try to guess if the images are generated or not. It has been used to assess generative models of images from ImageNet \cite{Denton15} and also on digit images \cite{Lake15}. Others propose to replace the human analysis by the analysis of first and second moments of activation of  a neural network. This method has been used with the output of the inception model for ``Inception Score'' (IS) \cite{Salimans16}, or with activation from intermediate layers for ``Frechet Inception Score'' (FID) \cite{Heusel17}. Log-likelihood based evaluation metrics were also widely used to evaluate generative models but as shown in \cite{Theis15}, those evaluations can be misleading in high dimensional cases such as images.

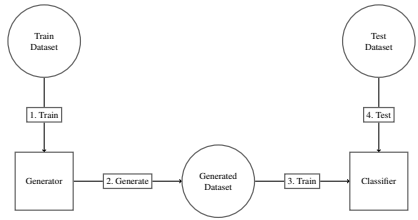
\begin{figure}
\begin{center}
\resizebox{0.45\textwidth}{!}{
\begin{tikzpicture}[scale=0.50,
roundnode/.style={circle, draw=black!60, fill=green!0, very thick, minimum size=25mm},
squarednode/.style={rectangle, draw=black!60, fill=red!0, very thick, minimum size=20mm},
squarednode2/.style={rectangle, draw=black!60, fill=red!0, very thick, minimum size=5mm},
]

\node[squarednode]      (generator)                              {Generator};
\node[squarednode2]      (trainOn)       [above=of generator] {1. Train};
\node[roundnode]        (train)       [above=of trainOn, align=center] {Train \\ Dataset};

\node[squarednode2]        (generate)       [right=of generator] {2. Generate}; 
\node[roundnode]        (generated)       [right=of generate, align=center] {Generated \\ Dataset}; 

\node[squarednode2]      (trainOn2)       [right=of generated] {3. Train};

\node[squarednode]      (classifier)       [right=of trainOn2] {Classifier};
\node[squarednode2]      (testOn)       [above=of classifier] {4. Test};
\node[roundnode]        (test)       [above=of testOn, align=center] {Test \\ Dataset};
 
\draw[-] (train.south) -- (trainOn.north);
\draw[->] (trainOn.south) -- (generator.north);
\draw[-] (generator.east) -- (generate.west);
\draw[->] (generate.east) -- (generated.west);
\draw[-] (generated.east) -- (trainOn2.west);
\draw[->] (trainOn2.east) -- (classifier.west);
\draw[-] (test.south) -- (testOn.north);
\draw[->] (testOn.south) -- (classifier.north);

\end{tikzpicture}

}
\label{fig:shema_methode}
\caption{Proposed method : 1. Train a generator on real training data, 2. Generate labeled data, 3. Train classifier with the generated data, 4. Evaluate the generator by testing the classifier on the test set composed of real data}
\label{fig:shema_methode}
\end{center}
\end{figure}

The solution we propose evaluates generative models by training a classifier with generated samples and testing it on real data.
The classifier test accuracy indicates how good the generative model is at reproducing the distribution of the original dataset. The full process of the method is illustrated in Figure \ref{fig:shema_methode}.  
The test data come from the original dataset but have not been used to train the generator.

Our contribution is twofolds: first we propose a method to evaluate generative models on the testing set.  
Secondly we introduce a quantitative score, \textit{the fitting capacity} (FiC), to evaluate and compare performance of generative models.

\section{Related work}
\label{Related work}
\subsection{Generative models}

Generative models can be implemented in various frameworks and settings. In this paper we focus on two kind of those frameworks : variational auto-encoders and generative adversarial networks.

The variational auto-encoder (VAE) framework \cite{Kingma13}, \cite{Rezende14} is a particular kind of auto-encoder which learns to map data into a Gaussian latent space, generally chosen as an univariate normal distribution $\mathbf{N}(0,I)$ (where $I$ is the identity matrix). The VAE learns also the inverse mapping from this  latent space to the observation space. This characteristic makes the VAE an interesting option for generating new data after training. Indeed, thanks to the inverse mapping, new data can be generated by sampling a Gaussian distribution and decoding these samples.  The particularity of the latent space comes from the minimization of the KL divergence between the distribution of data in the latent space and the prior $\mathbf{N}(0,I)$. For the sake of simplicity, in this paper we will refer to decoder of the VAE as a generator.

Generative adversarial networks \cite{Goodfellow14} are another framework of models that learn to generate data. The learning process is a game between two networks: a generator that learns to produce images from a distribution $P$ and a discriminator which learns to discriminate between generated and true images. The generator learns to fool the discriminator and the discriminator learns to not be fooled. This class of generative models can produce visually realistic samples from diverse datasets but they suffer from instabilities in their training.  One of the models we evaluate, the Wasserstein GAN (WGAN) \cite{Arjovsky17}, try to address those issues by enforcing a Lipschitz constraint on the discriminator. We also evaluate the BEGAN \cite{Berthelot17}, another variant of the GAN which uses an auto-encoder as discriminator.

Both GANs and VAEs can also be implemented into a conditional setting.
Conditional neural networks \cite{Sohn15} and in particular Conditional Variational Autoencoders (CVAE) and Conditional Generative adversarial networks (CGAN) \cite{Mirza14} are a class of generative models that have control over the sample's class of their training dataset. By imposing a label during training on the generator, a conditional generative network can generate from any class and thus produces labeled data automatically. The conditional approach has been used to improve the quality of generative models and make them more discriminative \cite{Odena16}. They are particularly adapted for our setup because we need to generate labeled datasets to train our classifiers.

\subsection{Evaluation of generated samples}

The evaluation of generative models has been addressed in various settings. 
\cite{Theis15} show that different metrics (such as Parzen windows, Nearest Neighbor or Log likelihood) applied to generative models can lead to different results. Good results in one application of a generative model can not be used as evidence of good performance in another application. Their conclusion is that evaluation based on sample visual quality is a bad indicator for the entropy of samples. Conversely, the log-likelihood can be used to produce samples with high entropy but it does not assure good visual quality. In this setting, high entropy means high variability in the samples.

More methods exist to evaluate generative networks as described in \cite{Borji18}. In particular, approaches that use a classifier trained on real data to evaluate generative models, \cite{Zhang16,Isola16}. \cite{lopez2016revisiting} proposes to use the two samples test with a classifier to evaluate if generated samples and original one are from the same distribution. Since generative models often produce visual artifacts it may be to easy to discriminate real images from generated one. 
Different quantitative evaluation have also been experimented in \cite{Jiwoong18} which compares models of GANs in various settings.  
These  quantitative evaluations are based on divergence or distances between real data or real features and generated ones. In our method we don't train the classifier with real data but with generated one.

\subsection{Multi-scale structural similarity}

Multi-scale structural similarity (MS-SIM, \cite{Wang03}) is a measurement that gives a way to incorporate image details at different resolutions in order to compare two images. This similarity is generally used in the context of image compression to compare images before and after compression. \\
Odena et al. \cite{Odena16} use this similarity to estimate the variability inside a class. They randomly sample two images of a class and measure the MS-SIM. If the value is high, then images are considered different. By operating this process on multiple data points $X$ of the class $Y$, the similarity gives an insight on the entropy of $P(X|Y)$: if the MS-SIM gives high result, the entropy is high (i.e. variate images); otherwise, the entropy is low. A good generator produces variate classes with variate images. MS-SIM is able to estimate the variability of the generated samples, however, it can not estimate if the sample comes from one or several modes of the distribution $P(X|Y)$. For example, if we want to generate images of cats, the MS-SIM similarity can not differentiate a generator that produces different kinds of black cats from a network that produces different cats of different colors. In our method, if the generator is able to generate in only one mode of the distribution $P(X|Y)$, the score will be low in the testing phase.

\subsection{Inception score}

One of the most used approach to evaluate a generative model is Inception Score (IS) \cite{Salimans16,Odena16}. The authors use an inception classifier model pretrained on ImageNet to evaluate the sample distribution.
They compute the conditional classes distribution at each generated sample $x$, $P(Y | X=x)$ and  the general classes distribution $P(Y)$ over the generated dataset.

They proposed the following score :
\begin{equation}
IS(X)=\exp(\mathbb{E}_X [KL (P(Y | X) \parallel P(Y))]  \mbox{ ,}
\label{inception_score}
\end{equation}

Where KL is the Kullback-Leibler divergence. The KL term can be rewritten :

\begin{equation}
  KL (P(Y | X) \parallel P(Y)) = H(P(Y | X), P(Y)) - H(P(Y|X)) \mbox{ ,}
\label{inception_score}
\end{equation}
Where $H(P(Y|X))$ is the entropy of $P(Y|X)$ and $H(P(Y | X), P(Y))$ the cross-entropy between $P(Y | X)$ and $P(Y)$. The entropy term is low when predictions given by the inception model have high confidence in one class only. The entropy term is high in other cases. The cross-entropy term is low when predictions given by the inception model gives unbalanced classes in the whole dataset and is high if the dataset is balanced. 

Hence, the inception score promotes when the inception model predictions have high confidence in varied classes. The hypothesis is that if the inception has high confidence in its prediction the image should look real.

Unfortunately, it does not estimate if the samples have intra-class variability (it does not take the entropy of $P(X|Y)$ into account). Hence, a generator that could generate only one sample per class with high quality would maximize the score.

One important restriction of IS is that the generative models to evaluate should produce images in ImageNet classes because the model required for the evaluation is pretrained on it and can not evaluate other classes.

\subsection{Frechet Inception Distance}

Another recent approach to evaluate generative adversarial networks is the Frechet Inception Distance (FID)~\cite{Heusel17}. The FID, as the inception score, is based on low moment analysis. It compares the mean and the covariance of activations between real data ($\mu$ and $C$) and generated data ($\mu_{gen}$, $C_{gen}$). The activation are taken from a inner layer in an inception model. The comparison is done using the Frechet distance (as if the means and covariance where taken from a Gaussian distribution) (see Eq. \ref{eq:frechet_inception_distance}). The inception model is trained on Imagenet.

\begin{multline}
d^2((\mu,C),(\mu_{gen},C_{gen})) =\parallel \mu - \mu_{gen} \parallel_2^2
+ Tr(C+C_{gen} -  2(C*C_{gen})^{\frac{1}{2}})  \mbox{ ,}
\label{eq:frechet_inception_distance}
\end{multline}

FID have in particular been used in a large scale study to compare generative models \cite{Lucic17}. FID measure the similarities between the distribution of generated feature and the distribution of real features but it does not directly assess the good visual quality. Furthermore, it assumes a gaussian distribution of features over the dataset which introduce a bias in the evaluation.

Our approach is similar to the approach developed in parallel by \cite{Santurkar18}. However we evaluate all generative models with the same generator architecture, all trained by the same method and not models with their original architecture and training process. We have then a clear comparison of different learning criterion.

\section{Methods}
\label{sec:methods}
We evaluate generators in a supervised training setup. The real dataset $D$, the \textit{original dataset}, is composed of pairs of examples $(x, y)$ where $x$ is a data point and $y$ the associated label. The dataset is split in three subsets $D_{train}, D_{valid}$ and $D_{test}$ for cross-validation. 
Our method needs a generative model that can sample conditionally from any given label $y$, i.e. we can choose from which class we sample the generator. This conditional generative model is trained on $D_{train}$. Then, we sample the generator to crate a new dataset $D_{gen}$ of samples $\hat{x}$. $D_{gen}$ is sampled with balanced classes. It is used afterwards to train a classifier $C$ implemented as a deep neural network. $D_{valid}$ is used for validation of $C$. $C$ will then be used to evaluate $G$.

When $C$ is trained, we evaluate $C$ on $D_{test}$ the accuracy of $C$ on $D_{test}$ is what we call the ''fitting capacity'' (FiC). 

It can be compared with the accuracy of a classifier trained only on real data from $D_{train}$ (the baseline) or with the FiC of another generator. We consider that the baseline is an upperbound to the FiC because no generative models beats it when $\tau=1$. But it is theoretically possible to have a FiC higher to the baseline.

We can summarize our method as follows:
\begin{enumerate}
\item Train a conditional generative model over $D_{train}$
\item Sample data to produce $D_{gen}$
\item Train a discriminative model (the classifier) over $D_{gen}$
\item Select a classifier over a validation set $D_{valid}$.
\item Iterate the process for several generative models and compare the accuracy of the classifiers on $D_{test}$.
\end{enumerate}

The protocol presented allows to analyze the performance of a model on the whole test set or class by class. 
The final accuracy can be compared directly to with the accuracy compute with another generator even if the classifier change.
As we will see in results, we estimate the stability of the generators by training them with different random seeds. 

The simple act of changing these seeds can have great impact on the generative models training and thus induce a variability in the results. To show that the variability of results comes mainly from the instability of the generator and not from the classifier, we compare our results with results computed with KNN classifiers instead of neural networks. As KNN classifiers are deterministic, if the random seeds produce variability with this kind of classifier,   the instability necessarily comes from the generators. The KNN classifier is however not a good option for our evaluation methods because it is not adapted for complex image classification.

The classifier was chosen to be a deep neural network because they are known to be difficult to train if the testing set distribution is biased with respect to the training set distribution. 
This characteristic is put to good use in order to compare generated datasets and hence generative models. If $D_{gen}$ contains unrealistic samples or samples that are not diverse enough, the classifier will not reach a high accuracy. Moreover, to investigate the impact of generated data into the training process of a classifier, we also experiment the method by mixing real data and generated data. The ratio between generated data over the complete dataset is called $\tau$. If $\tau=0$ there is no generated data and $\tau=1$ means only generated data. 

We call our final score for a generator $G$ the \textit{fitting capacity} (Noted $\Psi_G$) because it evaluates the ability of a generator to fit the distribution of a testing set. It is the testing accuracy of the classifier $C$ over $D_{Test}$ trained by a generator when $\tau=1$.

It is important to note that for fair comparison, whatever $\tau$ the total number of datapoint used to train the classifier is constant, i.e. more generated data lead to less real data.

We evaluate models with the generator or discriminator architecture proposed in \cite{Chen16}.

\section{Experiments}

\subsection{Implementation details}

Generative models are often evaluated on the MNIST dataset. Fashion MNIST \cite{Xiao2017} is a direct drop-in replacement for the original MNIST dataset but with more complex classes with higher variability. Thus, we use this dataset in order to evaluate different generative models in addition to MNIST.

As presented in the previous section, to apply our method, we need to generate labeled datasets. We used two different methods for that. For the first one, we train one generator for each class $y$ on $D_{train}$. This enables us to generate sample from a specific class and to generate a labeled dataset. In this setting, we compare VAE, WGAN, GAN and BEGAN. The second method uses conditional generative models which can generate the whole labeled dataset $D_{gen}$ directly.  In this case, we ran our experiments on CVAE and CGAN. The generators are trained with Adam optimizer on the whole original training dataset for $25$ epochs on eight different random seeds.

The classifier model trained on $D_{gen}$ is a standard deep CNN with a softmax output (see in Appendix for details).
The classifier is trained with Adam optimizer for a maximum of $200$ epochs. We use early stopping to stop the training if the accuracy does not improve anymore on the validation set after $50$ epochs. Then, we apply model selection based on the validation error and compute the test error on $D_{test}$. The architecture of the classifier is kept fixed for all experiments.

We performed experiment with values of  $\tau = [0.0, 0.125, 0.250,..., 1.0]$. $\tau = 0$  is used as a baseline (using only real data) to compare other results. However most of the analysis are made on $\tau=1$ because the results are representative of the full quality of the generator, i.e. generalization and fitting of the distribution of the testing set. 

The experiment with various $\tau$ makes it possible to visualize if a generator is able to generalize and fit a given dataset. If the method results improve when $\tau$ is low ($\tau< 0.5$) it means that the generator is able to perform data augmentation i.e. generalize training data. If the result are as good when $\tau$ is high ($\tau > 0.5$) and a fortiori near 1 it means the generator is able to fit $D_{test}$ distribution.

\subsection{Adaptation of Inception Score and Frechet Inception Distance}
We compare our evaluation method of the generative model with the two most used methods : IS and FID.
IS and FID, as originally defined, use a model pretrained on ImagenNet. To apply these methods, it is mandatory to use the exact same model proposed in \cite{Salimans16} with the exact same parameter because otherwise results are not comparable with results from other papers. However, as proposed in \cite{li17} we can adapt these methods to other setting (for us, MNIST and Fashion-MNIST) to compare several generative models with each other. 

We therefore train a model for classification on $D_{train}$. Then we use it to produce a probability vector to compute IS and an activation vector to compute FID. The activation are from a layer in the middle of the model (details in appendix).
The very same formula than IS and FID can then be used to compare models. 

\subsection{Results}

\begin{figure}[tb]

       \centering
    \begin{subfigure}[b]{0.35\textwidth}
        \includegraphics[width=\textwidth]{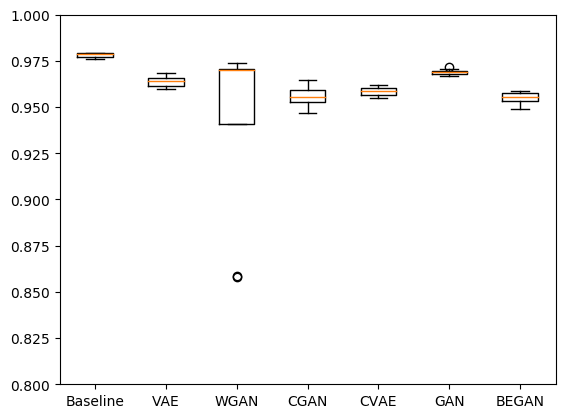}
        \caption{MNIST boxplot}
        \label{fig:diagram-MNIST}
    \end{subfigure}
   \begin{subfigure}[b]{0.35\textwidth}
       \includegraphics[width=\textwidth]{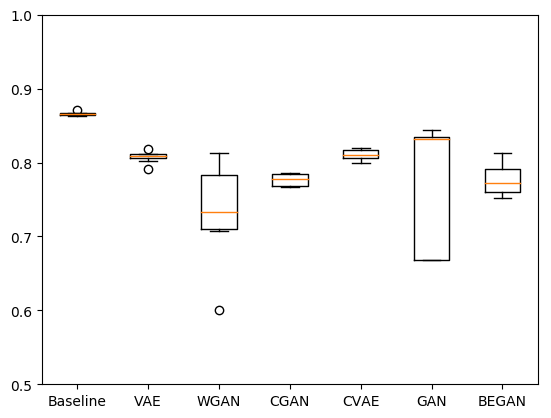}
        \caption{fashion-MNIST boxplot}
        \label{fig:diagram-fashion}
   \end{subfigure}
   \caption{Analysis and comparison of fitting capacity for different models when $\tau=1$}
    \label{fig:diagram}

\end{figure}   
   
The relevant results presented below are the maximum \textit{fitting capacity} (FiC) of each model over all seeds in order to evaluate what models can achieve and statistics on the results among those 8 seeds to give insight on the stability of each model with regards to the random seed.

First we present boxplots of \textit{FiC} results of each models in Figures \ref{fig:diagram-MNIST} and \ref{fig:diagram-fashion}. They present the median value along with the first and last quartile limits. Furthermore they display the outliers of each training (values that are outside 1.5 of the interquartile range (IQR) over the different seeds). This representation is less sensible to outliers than mean value and standard deviation without making those outliers disappear\footnote{The Figures \ref{fig:diagram-MNIST} and \ref{fig:diagram-fashion} are zoomed to be able to visually discriminates models making some outliers out of the plot. Full figures are in appendix.}. 

Those results show an advantage for the GAN model as it has the best median value (even if it does not make better than baseline). Unfortunately some of the generator training failed (in particular on Fashion-Mnist), producing outliers in the results. WGAN produce results comparable to GAN in MNIST but seems less stable. 

The figures are complemented with the values computed for the mean \textit{FiC} $\Psi_G$ in Table~\ref{table:Mean_Table} and the values for the best \textit{FiC} in Table~\ref{table:Best_Table}.
We can note that for both MNIST and Fashion-MNIST, models with unstable results, such as GAN and WGAN, have the best $\Psi_G$. However since some training failed for GAN and WGAN, more stable models, such as VAE and CGAN, have the best \textit{mean fitting capacity} (FiC).

\begin{table}[tbp]
\centering
\caption{Mean $\Psi_G$}
\label{table:Mean_Table}
\begin{adjustbox}{width=\textwidth}
\begin{tabular}{|l |l | l |l | l | l | l | l | l | l}
\hline
 Datasets    & \textbf{Baseline} &VAE       & CVAE      & GAN & CGAN & WGAN & BEGAN \\
 \hline
 MNIST       & \textbf{97.81\%}  & 96.39\% & 95.86\% & 86.03\% & \textbf{96.45\%} & 94.25\% & 95.49\%\\
 \hline
 Fashion   & \textbf{86.59\%} & \textbf{80.75\%} & 73.43\% & 67.75\% & 77.68\% & 73.43\% & 77.64\%  \\
 \hline
 
\end{tabular}
\end{adjustbox}
\end{table}

\begin{table}[tbp]
\centering
\caption{Best $\Psi_G$}
\label{table:Best_Table}
\begin{adjustbox}{width=\textwidth}
\begin{tabular}{|l |l | l |l | l | l | l | l | l | l}
\hline
 Datasets    & \textbf{Baseline} &VAE      & CVAE    & GAN     & CGAN    & WGAN & BEGAN \\
 \hline
 MNIST       & \textbf{97.94\%}  & 96.82\% & 96.21\% & 97.18\% & 96.45\% & \textbf{97.37\%} &  95.86\%\\
 \hline
 Fashion   & \textbf{87.08\%} & 81.85\% & 81.93\% & \textbf{84.43}\% & 78.63\% & 81.32\% & 81.32\%  \\
 \hline
 
\end{tabular}
\end{adjustbox}
\end{table}

\begin{figure}
    \centering
    \begin{subfigure}[b]{0.45\textwidth}
        \includegraphics[width=\textwidth]{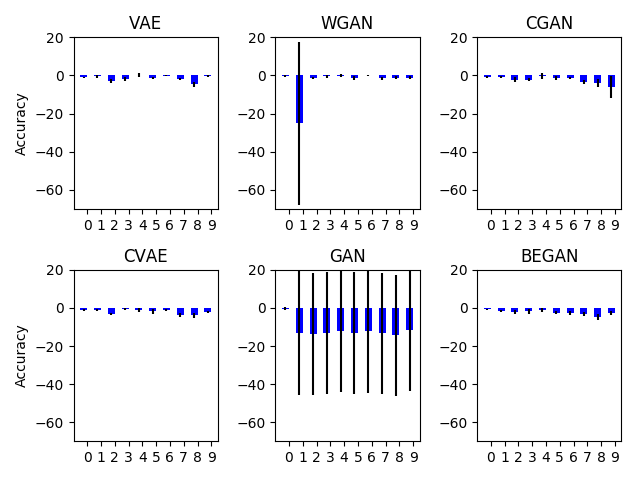}
        \caption{Relative accuracy wrt. baseline on mnist class by class for $\tau=1$}
        \label{fig:accuracy-MNIST_classes}
    \end{subfigure}
   \begin{subfigure}[b]{0.45\textwidth}
       \includegraphics[width=\textwidth]{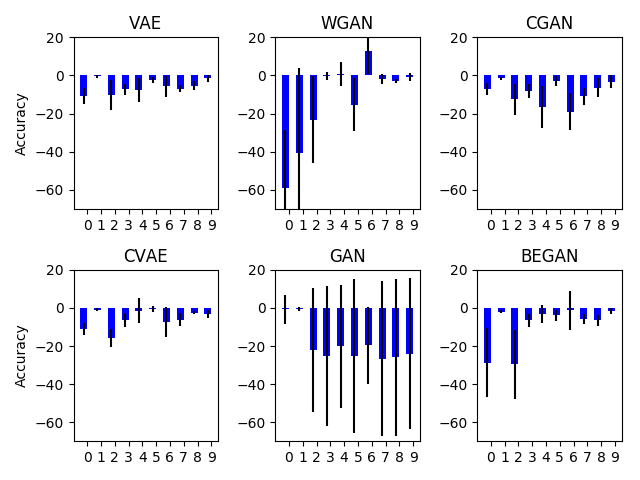}
        \caption{Relative accuracy wrt. baseline on fashion-mnist class by class for $\tau=1$}
        \label{fig:accuracy-fashion_classes}
   \end{subfigure}
   
   \caption{Plot of  the difference between models performance and baseline class by class when $\tau=1$ : Mean and standard deviation over random seeds}
\end{figure} 

In addition, we present in Figures \ref{fig:accuracy-MNIST_classes} and \ref{fig:accuracy-fashion_classes}, the \textit{per class fitting capacity}. The figures show the difference between the baseline classes results and the classifier trained on the generator to evaluate.
For  generative models that are not conditional and trained class by class, those figures show how the generator is successful to generate in different classes of the dataset. For conditional generative models, it evaluates if the models is able to learn each mode of the distribution with the same accuracy.

We can also estimate the stability of each model class by class or from a class to another. For example we can see that WGAN on MNIST is very stable except on  class $1$ and on Fashion-MNIST it seems to struggle a lot between the first 3 classes. On an other hand we can see that BEGAN has some trouble on Fashion-MNIST on class $0$ and $2$ (T-shirt and pullover) suggesting that the generator is not good enough to discriminate between those two classes.

\begin{figure*} [t]  
       \centering
    \begin{subfigure}[b]{0.39\textwidth}
        \includegraphics[width=\textwidth]{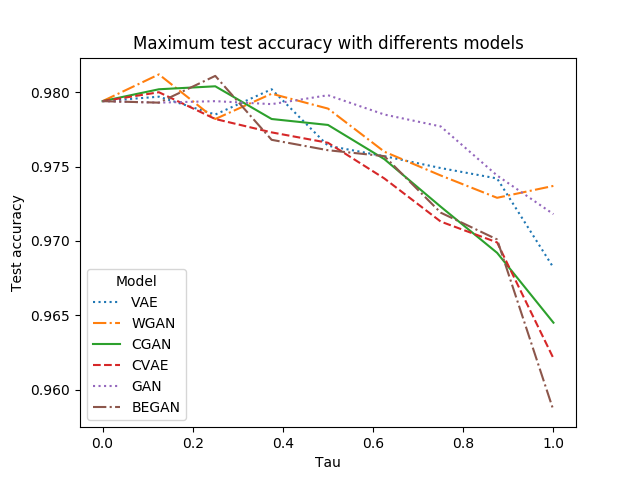}
        \label{fig:accuracy-MNIST}
    \end{subfigure}
   \begin{subfigure}[b]{0.35\textwidth}
       \includegraphics[width=\textwidth]{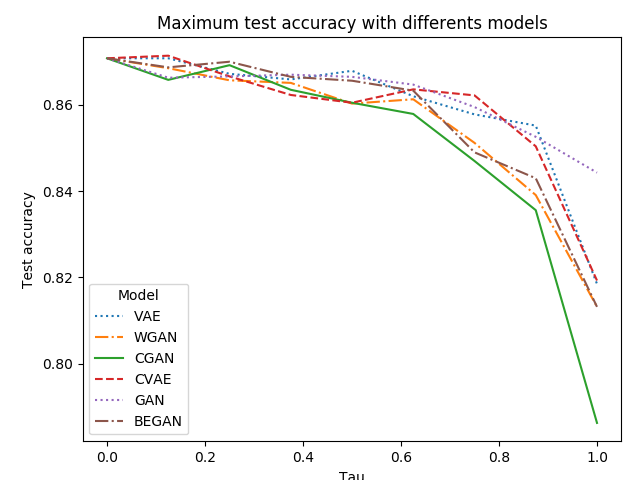}
        \label{fig:accuracy-fashion}
   \end{subfigure}
\caption{Representation of the  test accuracy of the classifiers trained by each $G$ with various $\tau$ on MNIST (left) and Fashion-MNIST (Right). Figures show maximum test accuracy for each models against $\tau$.}
\label{fig:accuracy}
\end{figure*}

\begin{figure*}
   
   \centering
    \begin{subfigure}[b]{0.4\textwidth}
        \label{fig:accuracy-MNIST_var}
        \includegraphics[width=\textwidth]{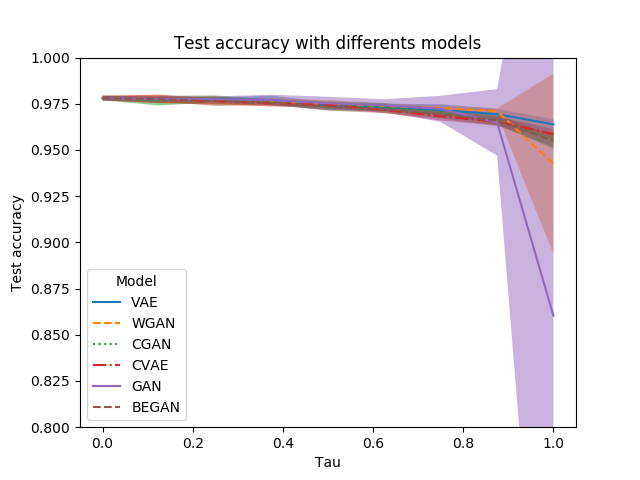}
    \end{subfigure}
   \begin{subfigure}[b]{0.35\textwidth}
       \includegraphics[width=\textwidth]{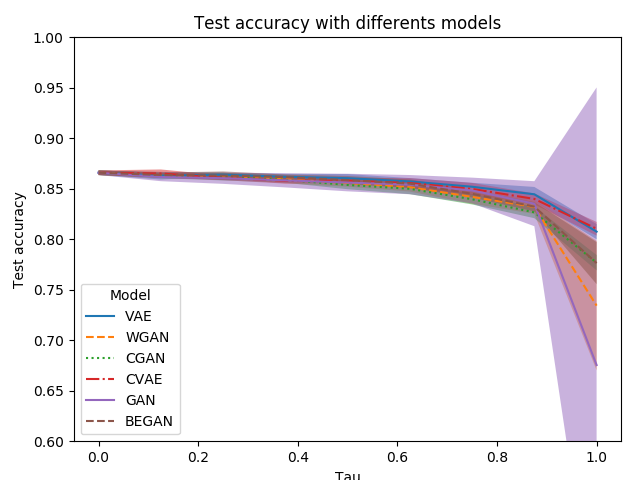}
        \label{fig:accuracy-fashion_var}
   \end{subfigure}
   
\caption{Representation of the  test accuracy of the classifiers trained by each $G$ with various $\tau$ on MNIST (left) and Fashion-MNIST (Right). Figures show mean and standard deviation of classifiers test accuracy}
\label{fig:accuracy_var}
\end{figure*}

In Figure {\ref{fig:accuracy}}, we present the test accuracy with various $\tau$ to represent the impact of generated data on the training process. When $\tau=0$, there is no generated data, this is the result of the baseline. Figure \ref{fig:accuracy} show result of cherry-picking among seeds the best results for each $\tau$, Figure \ref{fig:accuracy_var} show statistics among different seeds and the stability of each model on the trained dataset.
Our interpretation is that if the accuracy is better than baseline with a low $\tau$ ( $0< \tau < 0.5$) it means that the generator is able to generalize by learning meaningful information about the dataset. When $\tau > 0.5 $ if the accuracy is still near the baseline it means the generated data can replace the dataset in most parts of the distribution. When $\tau=1$, the classifier is trained only on generated samples. If the accuracy is still better than the baseline, it means that the generator has fitted the training distribution (and eventually has learned to generalize if this score is high over the test set).

Our results show that some models are able to produce data augmentation and outperform the baseline when $\tau$ is low but unfortunately none of them is able to do the same when $\tau$ is high. This show that they are not able to replace completely the true data in this setting. Following this interpretation, Figure \ref{fig:accuracy} allows us to compare different generative neural networks on both datasets. For example, we can see that all models expectation are equivalent when $\tau$ is low. However when $\tau$ is high we can clearly differentiate generative models type.

\begin{figure*}   
       \centering
    \begin{subfigure}[b]{0.4\textwidth}
        \includegraphics[width=\textwidth]{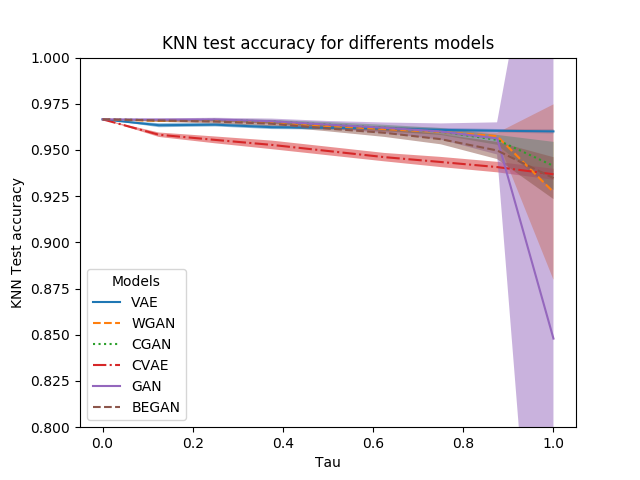}
        \label{fig:1nn_accuracy-MNIST}
    \end{subfigure}
   \begin{subfigure}[b]{0.4\textwidth}
       \includegraphics[width=\textwidth]{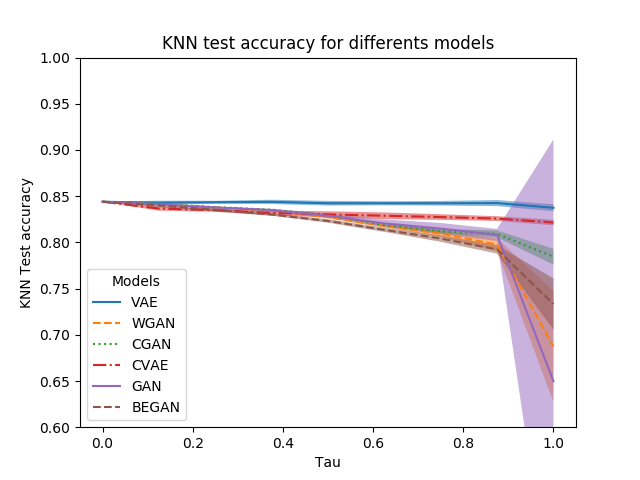}
        \label{fig:1nn_accuracy-fashion}
   \end{subfigure}
   \caption{Comparison of models using a nearest neighbor classifier on MNIST (left) and Fashion-MNIST (Right). Figures show mean and standard deviation of Classifiers Accuracy for 1-NN }
    \label{fig:1nn_accuracy}

\end{figure*}

Some of the curves in Figure \ref{fig:accuracy_var} have high standard deviation (e.g. for GAN when $\tau=1$). To show that it is not due to the classifier instability we plot the results of classification with various $\tau$ with a KNN classifier (Figure \ref{fig:1nn_accuracy} (k=1) ). KNN algorithms are stable since they are deterministic. The standard deviations found with KNN classifiers is similar to those with neural networks classifiers. This proves that the instability does not come from the classifier but from the generative models. This is coherent with the fact that in Figure \ref{fig:diagram}, the diagrams of the reference classifiers trained with true data on eight different seeds show a high stability of the classifier model.

\subsection{Comparison with IS and FID}

We compare our results to IS and FID methods. The two methods have  been slightly adapted to fit our setting as described previously in Section \ref{sec:methods}.
To be able to compare easily the different methods, we normalize values in order to have a mean of $0$ and a standard deviation of $1$ among the different models. 
Originally for FID methods, a low value means a better model, for easier comparison we multiply the FID score by $-1$ to valuate by an higher value a better model as for the other evaluation methods.
The results are shown on Figure \ref{fig:Comparison}. 
We added a baseline for each method, for the \textit{fitting capacity} the baseline is the test accuracy when the classifier trained on true data, the inception score baseline is computed on the test data and the frechet inception distance (FID) baseline is computed between train data and test data.

The results of the IS are completely different between MNIST and Fashion-MNIST and some model radically beat the baseline (as WGAN) when in the other methods, none of the results outperforms  the baselines. 
However, the FID computed between test set and generated data gives coherent results between MNIST and fashion-MNIST. Even if they are not always coherent with our results. As an example, VAE does not perform well with FID when we can see with our \textit{fitting capacity} that it is able to train a classifier quite well with high stability. The FID baseline outperform the other models however there is a small margin between best model and baseline when the margin between best model and baseline is big for the \textit{fitting capacity}.
The performances of each model are unfortunatly specific to each datasets and the experiments made are not sufficient to generalized results to other datasets. This can be seen in models like CGAN or WGAN where results are very different between MNIST and Fashion-MNIST.

\section{Discussion}

We presented a method to evaluate a generative model: the \textit{fitting capacity}. It assesses how well the generative model learned to generalize and fit a distribution in a conditional setting. Moreover, it gives a clear insight on potential top-down application of generative models.

The use of a discriminative model to evaluate generative models have already be experimented, e.g with IS and FID. However, the model used in those methods is pretrained on true data and can not be replaced with another one. In other case, results are not comparable.
Fitting capacity is based on the testing set, which is by definition specifically designed to measure the ability to understand and generalize data distribution. Therefor, it is well adapted to evaluate a learning algorithm as a generative model. 
On the other hand, relying only on the testing set, make easy comparison between different approaches. Any generators with any classifiers can be compared as long as they evaluate on the same test set.

Moreover, \textit{fitting capacity} gives a more in depth evaluation of generated data. Indeed, IS and FID rely on the analysis of first and second moments of features when \textit{Fitting capacity} rely on data complex hidden variables : labels. Evaluating the joint distribution between labels and images into generated data gives a good insight on if the generative model fits the images distribution.

Our evaluation method requires more computational power with respect to other methods because a classifier needs to be trained. Unfortunately, This computational power, is unfortunately needed to evaluate if the data distribution is well fitted by the generative model. 
Nevertheless, we can take advantage from simpler approaches as IS and FID for model selection during training and then applying \textit{fitting capacity} for deeper analysis.

Our experiments are restricted to two simple datasets (MNIST and Fashion-MNIST), however our results show a instability for certain models with regards to the random seed. Therefore, results should be interpreted carefully as different initialization could gives different results. In order to disentangle what models can do versus what they actually do, we evaluate and compare generative models on their top performance and on their stability with regards to random seeds. 

In order to have a fair comparison we use same classifier to evaluate all generative models.
However, any classifier could be used in replacement to improve and complete the results, our focus is on keeping the same test set to compare models. For instance it is obvious that a generator that just learned to reproduce the training set will beat our results, since no generator are able to beat the baseline. However we hope that new generators or better classifier will be able to beat the baseline. 
We believe that it is simple enough to be easily reproducible to evaluate any generative model.  Moreover, \textit{fitting capacity} is well adapted to be used on other kinds of generative models that produce data that are not images.

\section{Conclusion}
This paper introduces a method to assess and compare the performances of generative models by training a classifier on generated samples. It estimate the ability of a generative model to fit and generalize a testing set. It does not directly assess the realistic characteristics of the generated data but rather if their content and variability contains enough information to classify real data.

This method makes it possible to take into account complex characteristics of the generated samples and not only the distribution of their features. Moreover it does not evaluate generative models by testing if we can discriminate true data from generated one. 
Our results show that the \textit{fitting capacity} allows to compare easily generative models and estimate their stability and shortcomings. It makes possible to both evaluate on the full dataset and evaluate class by class.
In our experiment, the fitting capacity suggest that to get the best results GAN or WGAN approach should be privileged, however to maximize the chance of having a decent result CGAN or VAE are preferable. 

We believe that generating data might have numerous application in top down settings: e.g. data-augmentation, data compression, transfer learning … We hope that this evaluation will help to select efficient and powerful generative models for those applications.

%
%
%
\bibliographystyle{splncs04}
\bibliography{bibliography.bib}

\begin{thebibliography}{10}
\providecommand{\url}[1]{\texttt{#1}}
\providecommand{\urlprefix}{URL }
\providecommand{\doi}[1]{https://doi.org/#1}

\bibitem{Arjovsky17}
Arjovsky, M., Chintala, S., Bottou, L.: {W}asserstein generative adversarial
  networks. In: Proceedings of the 34th International Conference on Machine
  Learning (2017)

\bibitem{Berthelot17}
Berthelot, D., Schumm, T., Metz, L.: {BEGAN:} boundary equilibrium generative
  adversarial networks. CoRR  \textbf{abs/1703.10717} (2017),
  \url{http://arxiv.org/abs/1703.10717}

\bibitem{Borji18}
{Borji}, A.: {Pros and Cons of GAN Evaluation Measures}. ArXiv e-prints  (Feb
  2018)

\bibitem{Chen16}
Chen, X., Duan, Y., Houthooft, R., Schulman, J., Sutskever, I., Abbeel, P.:
  Infogan: Interpretable representation learning by information maximizing
  generative adversarial nets. CoRR  \textbf{abs/1606.03657} (2016),
  \url{http://arxiv.org/abs/1606.03657}

\bibitem{Denton15}
Denton, E.L., Chintala, S., szlam, a., Fergus, R.: Deep generative image models
  using a laplacian pyramid of adversarial networks. In: Advances in Neural
  Information Processing Systems 28 (2015)

\bibitem{Goodfellow14}
Goodfellow, I., Pouget-Abadie, J., Mirza, M., Xu, B., Warde-Farley, D., Ozair,
  S., Courville, A., Bengio, Y.: Generative adversarial nets. In: Advances in
  Neural Information Processing Systems 27 (2014)

\bibitem{Heusel17}
{Heusel}, M., {Ramsauer}, H., {Unterthiner}, T., {Nessler}, B., {Hochreiter},
  S.: {GANs Trained by a Two Time-Scale Update Rule Converge to a Local Nash
  Equilibrium}. ArXiv e-prints  (Jun 2017)

\bibitem{Isola16}
Isola, P., Zhu, J., Zhou, T., Efros, A.A.: Image-to-image translation with
  conditional adversarial networks. CoRR  \textbf{abs/1611.07004} (2016),
  \url{http://arxiv.org/abs/1611.07004}

\bibitem{Jiwoong18}
{Jiwoong Im}, D., {Ma}, H., {Taylor}, G., {Branson}, K.: {Quantitatively
  Evaluating GANs With Divergences Proposed for Training}. ArXiv e-prints  (Mar
  2018)

\bibitem{Kingma13}
Kingma, D.P., Welling, M.: Auto-encoding variational bayes. In: Proceedings of
  the 2nd International Conference on Learning Representations (ICLR) (2014)

\bibitem{Lake15}
Lake, B.M., Salakhutdinov, R., Tenenbaum, J.B.: Human-level concept learning
  through probabilistic program induction. Science  (2015)

\bibitem{li17}
Li, C., Liu, H., Chen, C., Pu, Y., Chen, L., Henao, R., Carin, L.: Alice:
  Towards understanding adversarial learning for joint distribution matching.
  Neural Information Processing Systems (NIPS)  (2017)

\bibitem{lopez2016revisiting}
Lopez-Paz, D., Oquab, M.: Revisiting classifier two-sample tests. arXiv
  preprint arXiv:1610.06545  (2016)

\bibitem{Lucic17}
{Lucic}, M., {Kurach}, K., {Michalski}, M., {Gelly}, S., {Bousquet}, O.: {Are
  GANs Created Equal? A Large-Scale Study}. ArXiv e-prints  (Nov 2017)

\bibitem{Mirza14}
Mirza, M., Osindero, S.: Conditional generative adversarial nets. CoRR
  \textbf{abs/1411.1784} (2014)

\bibitem{ng2002discriminative}
Ng, A.Y., Jordan, M.I.: On discriminative vs. generative classifiers: A
  comparison of logistic regression and naive bayes. In: Advances in neural
  information processing systems (2002)

\bibitem{Nguyen16}
Nguyen, A., Clune, J., Bengio, Y., Dosovitskiy, A., Yosinski, J.: Plug \& play
  generative networks: Conditional iterative generation of images in latent
  space. In: Proceedings of the IEEE Conference on Computer Vision and Pattern
  Recognition (2017)

\bibitem{Odena16}
Odena, A., Olah, C., Shlens, J.: Conditional image synthesis with auxiliary
  classifier {GAN}s. In: Proceedings of the 34th International Conference on
  Machine Learning (2017)

\bibitem{Radford15}
Radford, A., Metz, L., Chintala, S.: Unsupervised representation learning with
  deep convolutional generative adversarial networks. CoRR
  \textbf{abs/1511.06434} (2015)

\bibitem{Ratner17}
Ratner, A.J., Ehrenberg, H.R., Hussain, Z., Dunnmon, J., R{\'e}, C.: Learning
  to compose domain-specific transformations for data augmentation. stat
  (2017)

\bibitem{Rezende14}
Rezende, D.J., Mohamed, S., Wierstra, D.: Stochastic backpropagation and
  approximate inference in deep generative models. In: Proceedings of the 31th
  International Conference on Machine Learning, (2014)

\bibitem{Salimans16}
Salimans, T., Goodfellow, I.J., Zaremba, W., Cheung, V., Radford, A., Chen, X.:
  Improved techniques for training gans. CoRR  \textbf{abs/1606.03498} (2016)

\bibitem{Santurkar18}
Santurkar, S., Schmidt, L., Madry, A.: A classification-based perspective on
  {GAN} distributions (2018), \url{https://openreview.net/forum?id=S1FQEfZA-}

\bibitem{Sixt16}
Sixt, L., Wild, B., Landgraf, T.: Rendergan: Generating realistic labeled data.
  CoRR  \textbf{abs/1611.01331} (2016)

\bibitem{Sohn15}
Sohn, K., Lee, H., Yan, X.: Learning structured output representation using
  deep conditional generative models. In: Advances in Neural Information
  Processing Systems 28 (2015)

\bibitem{Theis15}
Theis, L., van~den Oord, A., Bethge, M.: A note on the evaluation of generative
  models. In: Proceedings of the 4th International Conference on Learning
  Representations (ICLR) (2016)

\bibitem{Wang17}
{Wang}, W., {Wang}, A., {Tamar}, A., {Chen}, X., {Abbeel}, P.: {Safer
  Classification by Synthesis}. ArXiv e-prints  (Nov 2017)

\bibitem{Wang03}
Wang, Z., Simoncelli, E.P., Bovik, A.C.: Multiscale structural similarity for
  image quality assessment. In: Proc 37th Asilomar Conf on Signals, Systems and
  Computers (2003)

\bibitem{Xiao2017}
Xiao, H., Rasul, K., Vollgraf, R.: Fashion-mnist: a novel image dataset for
  benchmarking machine learning algorithms (2017)

\bibitem{Zhang16}
{Zhang}, R., {Isola}, P., {Efros}, A.A.: {Colorful Image Colorization}. ArXiv
  e-prints  (Mar 2016)

\end{thebibliography}
\appendix
\section*{APPENDIX}

\section{models}

\subsection{Generator Architectures}

\begin{table}[!htbp] 
\caption{ VAE and GAN Generator Architecture}
\label{tab:metrics}
\begin{center}
\begin{small}
\begin{tabular}{|p{20mm}|p{60mm}|}
\hline
\textbf{Layer} &
\textbf{Architecture} \\\hline\hline

1 & FC (20,1024) + BN + relu\\\hline
2 & FC (1024,128*7*7)  + BN + relu\\\hline
3 & ConvTranspose2d(128, 64, 4, 2, 1)  + BN + relu\\\hline
4 & nn.ConvTranspose2d(64, 20, 4, 2, 1) + sigmoid \\\hline

\end{tabular}
\end{small}
\end{center}
\end{table}

\subsection{Classifier Architectures}

\begin{table}[!htbp] 
\caption{MNIST}
\label{tab:metrics}
\begin{center}
\begin{small}
\begin{tabular}{|p{20mm}|p{60mm}|}
\hline
\textbf{Layer} &
\textbf{Architecture} \\\hline\hline

1 & conv(5*5), 10 filters + maxpool(2*2) + relu\\\hline
2 & conv(5*5), 20 filters + maxpool(2*2) + relu \\\hline
3 & dropout(0.5) \\\hline
4 & FC (320, 50 ) + relu \\\hline
5 & FC (50, 10 ) + log-softmax  \\\hline

\end{tabular}
\end{small}
\end{center}
\end{table}

\begin{table}[!htbp] 
\caption{Fashion-MNIST}
\label{tab:metrics}
\begin{center}
\begin{small}
\begin{tabular}{|p{20mm}|p{60mm}|}
\hline
\textbf{Layer} &
\textbf{Architecture} \\\hline\hline

1 & conv(5*5), 16 filters + maxpool(2*2) + relu\\\hline
2 & conv(5*5), 32 filters + maxpool(2*2) + relu \\\hline
3 & dropout(0.5) \\\hline
4 & FC (512, 10 ) + log-softmax \\\hline

\end{tabular}
\end{small}
\end{center}
\end{table}

\textbf{FID Note :} The activation vector to compute FID have been taken arbitrary at he ouput of layer 2 for both MNIST and Fashion-MNIST

\begin{figure}   
       \centering
    \begin{subfigure}[b]{0.35\textwidth}
        \includegraphics[width=\textwidth]{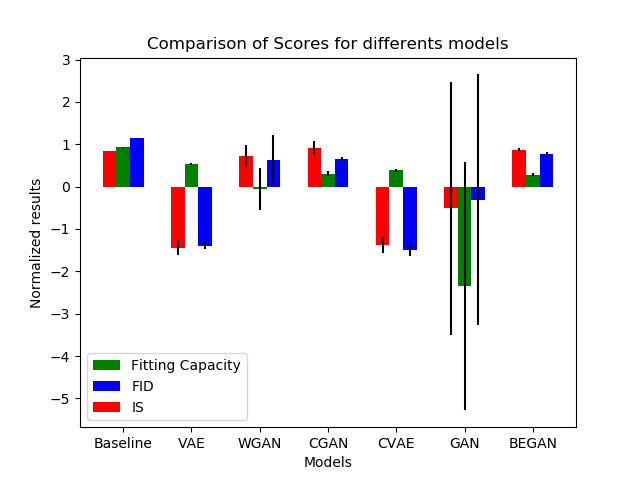}
        \label{fig:Comparison-MNIST}
    \end{subfigure}
   \begin{subfigure}[b]{0.35\textwidth}
       \includegraphics[width=\textwidth]{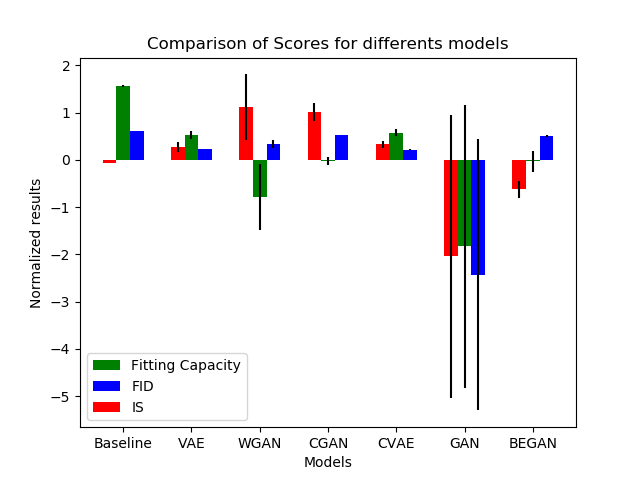}
        \label{fig:Comparison-fashion}
   \end{subfigure}
\caption{Comparison between results with mean and standard deviation over random seeds from different approaches on MNIST (left) and Fashion-MNIST  (right). Each result have been normalize to have mean=0 and standard deviation=1 among all models.}
        \label{fig:Comparison}
\end{figure}

\section*{Acknowledgement}
We really want to thanks Florian Bordes for experiment settings and interesting discussions as well as Pascal Vincent for his helpful advises. We would like also to thanks Natalia D\'iaz Rodriguez and Anthonin Raffin for their help in proof reading this article.
\appendix
\section*{APPENDIX}

\section{models}

\subsection{Generator Architectures}

\begin{table}[!htbp] 
\caption{ VAE and GAN Generator Architecture}
\label{tab:metrics}
\begin{center}
\begin{small}
\begin{tabular}{|p{20mm}|p{60mm}|}
\hline
\textbf{Layer} &
\textbf{Architecture} \\\hline\hline

1 & FC (20,1024) + BN + relu\\\hline
2 & FC (1024,128*7*7)  + BN + relu\\\hline
3 & ConvTranspose2d(128, 64, 4, 2, 1)  + BN + relu\\\hline
4 & nn.ConvTranspose2d(64, 20, 4, 2, 1) + sigmoid \\\hline

\end{tabular}
\end{small}
\end{center}
\end{table}

\subsection{Classifier Architectures}

\begin{table}[!htbp] 
\caption{MNIST}
\label{tab:metrics}
\begin{center}
\begin{small}
\begin{tabular}{|p{20mm}|p{60mm}|}
\hline
\textbf{Layer} &
\textbf{Architecture} \\\hline\hline

1 & conv(5*5), 10 filters + maxpool(2*2) + relu\\\hline
2 & conv(5*5), 20 filters + maxpool(2*2) + relu \\\hline
3 & dropout(0.5) \\\hline
4 & FC (320, 50 ) + relu \\\hline
5 & FC (50, 10 ) + log-softmax  \\\hline

\end{tabular}
\end{small}
\end{center}
\end{table}

\begin{table}[!htbp] 
\caption{Fashion-MNIST}
\label{tab:metrics}
\begin{center}
\begin{small}
\begin{tabular}{|p{20mm}|p{60mm}|}
\hline
\textbf{Layer} &
\textbf{Architecture} \\\hline\hline

1 & conv(5*5), 16 filters + maxpool(2*2) + relu\\\hline
2 & conv(5*5), 32 filters + maxpool(2*2) + relu \\\hline
3 & dropout(0.5) \\\hline
4 & FC (512, 10 ) + log-softmax \\\hline

\end{tabular}
\end{small}
\end{center}
\end{table}

\textbf{FID Note :} The activation vector to compute FID have been taken arbitrary at he ouput of layer 2 for both MNIST and Fashion-MNIST

\begin{figure}   
       \centering
    \begin{subfigure}[b]{0.35\textwidth}
        \includegraphics[width=\textwidth]{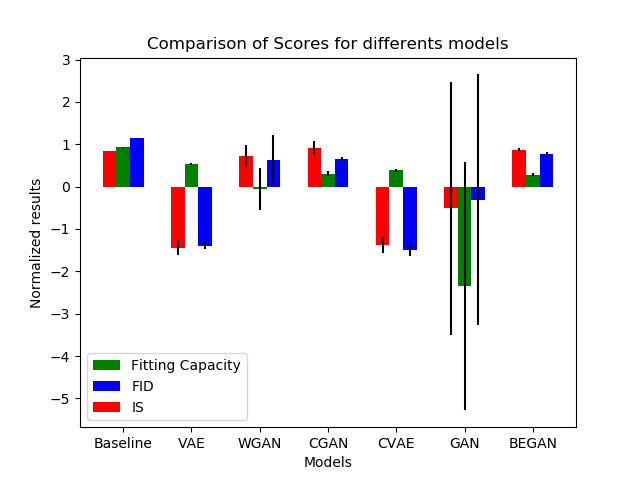}
        \label{fig:Comparison-MNIST}
    \end{subfigure}
   \begin{subfigure}[b]{0.35\textwidth}
       \includegraphics[width=\textwidth]{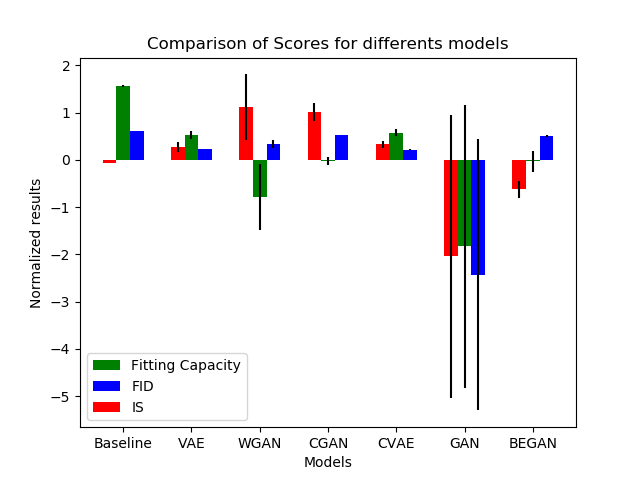}
        \label{fig:Comparison-fashion}
   \end{subfigure}
\caption{Comparison between results with mean and standard deviation over random seeds from different approaches on MNIST (left) and Fashion-MNIST  (right). Each result have been normalize to have mean=0 and standard deviation=1 among all models.}
        \label{fig:Comparison}
\end{figure}

\section*{Acknowledgement}
We really want to thanks Florian Bordes for experiment settings and interesting discussions as well as Pascal Vincent for his helpful advise. We would like also to thank Natalia D\'iaz Rodriguez and Anthonin Raffin for their help in proofreading this article.

\end{document}